\documentclass[letterpaper, 10 pt, conference]{ieeeconf}  

\IEEEoverridecommandlockouts                              

\usepackage{graphics} 
\usepackage{epsfig} 
\usepackage{mathptmx} 
\usepackage{multirow}
\usepackage[hyphens]{url}
\usepackage{hyperref}
\usepackage{graphicx}
\usepackage[table,xcdraw]{xcolor}
\usepackage{times} 
\usepackage{amsmath} 
\usepackage{amssymb}  
\usepackage{booktabs} 
\usepackage{graphicx}
\usepackage[table,xcdraw]{xcolor}
\newcommand{\method}{SwinDePose}

\begin{document}

\author{Zhujun Li$^{1}$ and Ioannis Stamos$^{2}$%
\thanks{$^{1}$Zhujun Li is computer science Ph.D. candidate at Graduate Center of the City University of New York., 365 5th Ave, New York, NY, USA
        {\tt\small zli3@gradcenter.cuny.edu}}%
\thanks{$^{2}$Ioannis Stamos is Professor of Computer Science at Hunter College and Graduate Center of 
        the City University of New York, 695 Park Avenue, New York, NY, USA
        {\tt\small istamos@hunter.cuny.edu}}%
}
\title{\LARGE \bf
Depth-based 6DoF Object Pose Estimation using Swin Transformer
}

\maketitle

\thispagestyle{empty}
\pagestyle{empty}

\begin{abstract}
Accurately estimating the 6D pose of objects is crucial for many applications, such as robotic grasping, autonomous driving, and augmented reality. However, this task becomes more challenging in poor lighting conditions or when dealing with textureless objects. To address this issue, depth images are becoming an increasingly popular choice due to their invariance to a scene's appearance and the implicit incorporation of essential geometric characteristics. However, fully leveraging depth information to improve the performance of pose estimation remains a difficult and under-investigated problem. To tackle this challenge, we propose a novel framework called {\method}, that uses only geometric information from depth images to achieve accurate 6D pose estimation. {\method} first calculates the angles between each normal vector defined in a depth image and the three coordinate axes in the camera coordinate system. The resulting angles are then formed into an image, which is encoded using Swin Transformer. Additionally, we apply RandLA-Net to learn the representations from point clouds. The resulting image and point clouds embeddings are concatenated and fed into a semantic segmentation module and a 3D keypoints localization module. Finally, we estimate 6D poses using a least-square fitting approach based on the target object's predicted semantic mask and 3D keypoints. In experiments on the LineMod and Occlusion LineMod datasets, SwinDePose outperforms existing state-of-the-art methods for 6D object pose estimation using depth images. This demonstrates the effectiveness of our approach and highlights its potential for improving performance in real-world scenarios. Our code is at \url{https://github.com/zhujunli1993/SwinDePose}.

\end{abstract}


\section{INTRODUCTION}
6D pose estimation involves determining the rigid transformation between camera coordinate and object coordinate systems, including the 3D rotation matrix and the 3D translation vector. This is a critical step in many significant applications, such as robotic manipulation \cite{tremblay2018deep, akinola2021dynamic}, autonomous driving \cite{wu20196d, sun2021rsn}, and augmented reality \cite{marchand2015pose, su2019deep}. For example, in robotic manipulation, robots need knowledge of the 6D poses of target objects for recognition and grasping \cite{collet2011moped}. In autonomous driving, vehicles must estimate the 6D poses of roads and obstacles for navigation \cite{alatise2020review}. In augmented reality, accurately estimating the 6D poses of real-world objects is essential for correctly placing virtual objects \cite{yu2005pose}.

6D pose estimation techniques are generally classified into three groups based on the type of input data: RGB input \cite{peng2019pvnet,lepetit2009epnp,yu20206dof}, RGB-D input \cite{wang2019densefusion,he2021ffb6d,he2020pvn3d}, and depth image input \cite{gao2021cloudaae,cai2022ove6d,gao20206d,liu2022catre}. RGB-based and RGB-D-based methods rely on the appearance information provided by RGB images, which limits their pose estimation performance in challenging scenarios, such as poor lighting or textureless objects. In contrast, depth images provide 3D geometric information that is less sensitive to lighting conditions or texture. Additionally, depth sensors have become more affordable, leading to a growing number of works that focus solely on using depth images for 6D pose estimation.

\begin{figure}[t] 
\includegraphics[width=0.5\textwidth]{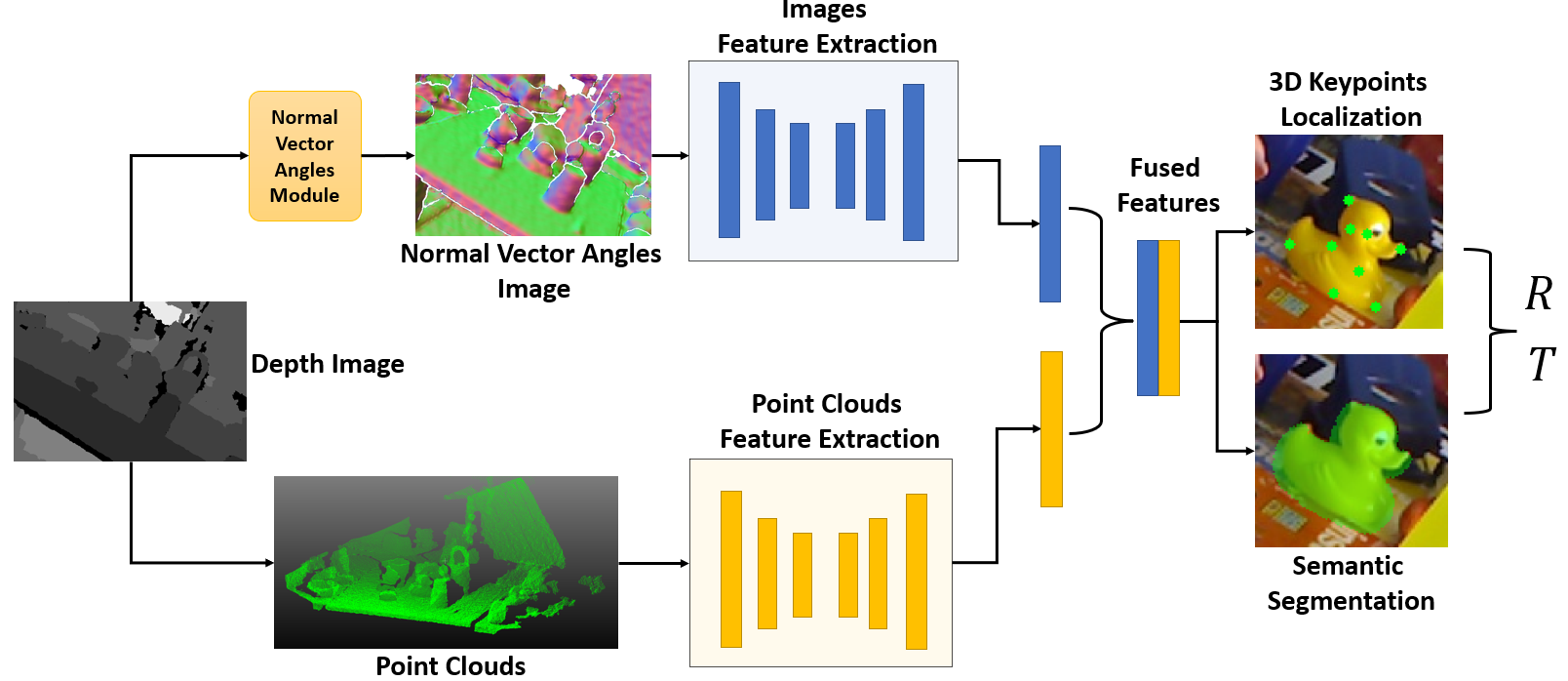}
\caption{We propose a novel framework called {\method} for 6D object pose estimation using depth images as input. Our approach lifts the depth image into point clouds and normal vector angles image, learns their representations for 3D keypoints localization and semantic segmentation, and finally recovers the 6D poses ($R$ and $T$) of the target object. In this figure, we project the localized 3D keypoints onto 2D pixels in the RGB image for better visualization.}
\label{first}
\end{figure}

Traditional depth-based 6D pose estimation methods, such as Point Pair Features (PPF) \cite{hinterstoisser2016going} and Fast Point Feature Histograms (FPFH) \cite{rusu2009fast}, rely on hand-crafted features based on objects' geometry information. However, these features are sensitive to changes in viewpoints, objects' shapes and appearances, and require significant human efforts for feature extraction and model fitting. Recent advancements in deep learning have led to the development of deep neural networks for learning geometric representations to estimate 6D poses from depth images \cite{gao2021cloudaae,gao20206d,liu2022catre}. These methods lift depth images to point clouds and design nerual networks for 3D geometry representation learning to estimate 6D poses. While these methods have shown promising results, they rely solely on point cloud embeddings and do not fully leverage the features in depth images. Therefore, incorporating 2D representations of depth images could provide additional information to improve pose estimation performance.

To introduce depth image embeddings into 6D pose estimation, treating a depth image as a gray-level intensity image and then applying existing vision algorithms are natural choices. However, traditional vision algorithms may not be well-suited for processing depth images due to noise, missing data, and differences in representation compared to RGB images. \cite{boubou2016differential} opened up new opportunities to overcome these challenges by using normal vectors at each surface point. In our work, we utilize normal vector angles to represent depth information. Specifically, we calculate the angles between each surface normal vector and the three coordinate axes. Then we normalize them to the RGB color range to form an RGB-like image, where each pixel has three normalized angle values. Such images are fed into our image representation learning network. In addition to the normal vector angles images, we lift depth images to point clouds using the given camera parameters and extract point clouds features via the point clouds representation learning network. Combining these two embeddings, our {\method} architecture can leverage depth images and point clouds information for more accurate 6D pose estimation.

Our proposed framework is shown in Fig. \ref{first}. We use Swin Transformer \cite{liu2021swin} for image feature extraction, while using RandLA-Net \cite{hu2020randla} for point cloud features extraction. The learned image and point cloud embeddings are then fused for 3D keypoints localization and semantic segmentation. Especially for image feature extraction, previous works \cite{he2021ffb6d,he2020pvn3d,wang2019densefusion} have used CNN-based networks. However, these methods have limitations such as small local receptive fields, sensitivity to object occlusions, and lack of global context. To overcome these limitations, we employ Swin Transformer, which leverages a self-attention mechanism to capture global context by attending to all positions in the input. Additionally, Swin Transformer exhibits robustness to deformations and occlusions, making it suitable for processing complex scenes. Works such as \cite{yuan2021hrformer,ma2022swinfusion} have applied Swin Transformer for computer vision tasks, such as object detection, instance semantic segmentation, and image classification. To the best of our knowledge, this is the first work using Swin Transformer for 6D object pose estimation based on depth information.

To evaluate our method, we conduct experiments on two popular datasets, the LineMod and Occlusion LineMod datasets. Experimental results show that the proposed approach outperforms the state-of-the-art depth-based methods.

To summarize, the main contributions of our work are:

\begin{itemize}
\item Generating normal vector angles images to fully leverage the geometry information from depth images, that can be combined with point clouds, for representation learning.
\item Introducing a novel framework including Swin Transformer and point cloud networks for 6D pose estimation.
\item Achieving state-of-the-art 6D pose estimation performance based on depth information on the LineMod and Occlusion LineMod datasets.
\end{itemize}

\section{RELATED WORKS}
\subsection{6D Pose Estimation from Depth Images.}
Most existing research on 6D pose estimation from depth images using deep learning has primarily focused on converting depth images into point clouds and utilizing existing semantic segmentation models to extract object masks from depth images. These masks are then used to crop objects from point clouds and feed them into their proposed 6D pose estimation framework. For instance, \cite{gao2021cloudaae} proposed a framework based on the augmented autoencoder, called CloudAAE, and trained it on a large synthetic point cloud dataset. \cite{cai2022ove6d} proposed the OVE6D method that was trained on a large synthetic image dataset containing ShapeNet objects. Other systems, such as those introduced by \cite{gao20206d} and \cite{liu2022catre}, utilized instance semantic segmentation masks from depth images and point clouds to regress 6D poses. In contrast, our proposed method includes a semantic segmentation module and integrates point clouds embeddings with normal vector angles images embeddings obtained from depth images to estimate 6D poses.

\subsection{Vision Transformer.}
The Transformer, originally developed for natural language processing tasks, has been adapted for computer vision tasks and has demonstrated significant improvements in performance. Researchers have designed various transformer networks for object detection, segmentation, and pose estimation tasks.  \cite{liu2021swin} proposed Swin Transformer, which constructs hierarchical feature maps from input images using shifted windows and has been applied as a backbone network for various tasks. For example, \cite{shen2022simcrosstrans} designed SimCrossTrans using Swin Transformer for 2D object detection task. \cite{yuan2021hrformer} proposed HRFormer using a multi-resolution parallel design with local-window self-attention for dense prediction. \cite{ma2022swinfusion} introduced a novel image fusion network for multi-modal image fusion and digital photography image fusion. Moreover, \cite{beedu2022video} utilized Swin Transformer to learn image representations for human pose estimation. Our framework uses Swin Transformer as an encoder to extract image features for 6D object pose estimation.

\subsection{Depth Images Feature.}
Exploring depth information from depth images has been a well-studied area in computer vision. For example, surface curvature \cite{vemuri1986curvature}, and kernel features based on depth images \cite{bo2011depth} have been used. Other studies \cite{boubou2016differential,zelener2016cnn} have computed normal vectors from depth images and used spherical angles to represent the normal vectors for object recognition. Inspired by these studies, we extract normal vectors from depth images and compute the angles between each normal vector and the $XYZ$ coordinate axes in the camera coordinate system. These angles are then normalized and grouped as an image, which is fed into the image features extraction module.

\section{METHOD}
Given the input images, 6D pose estimation predicts the rigid transformation of objects from the object coordinate system to the camera coordinate system. The 6D includes rotation matrix $R \in SO(3)$ and translation vector $T \in \mathbb{R}^3$.
\begin{figure*} 
\includegraphics[width=\textwidth]{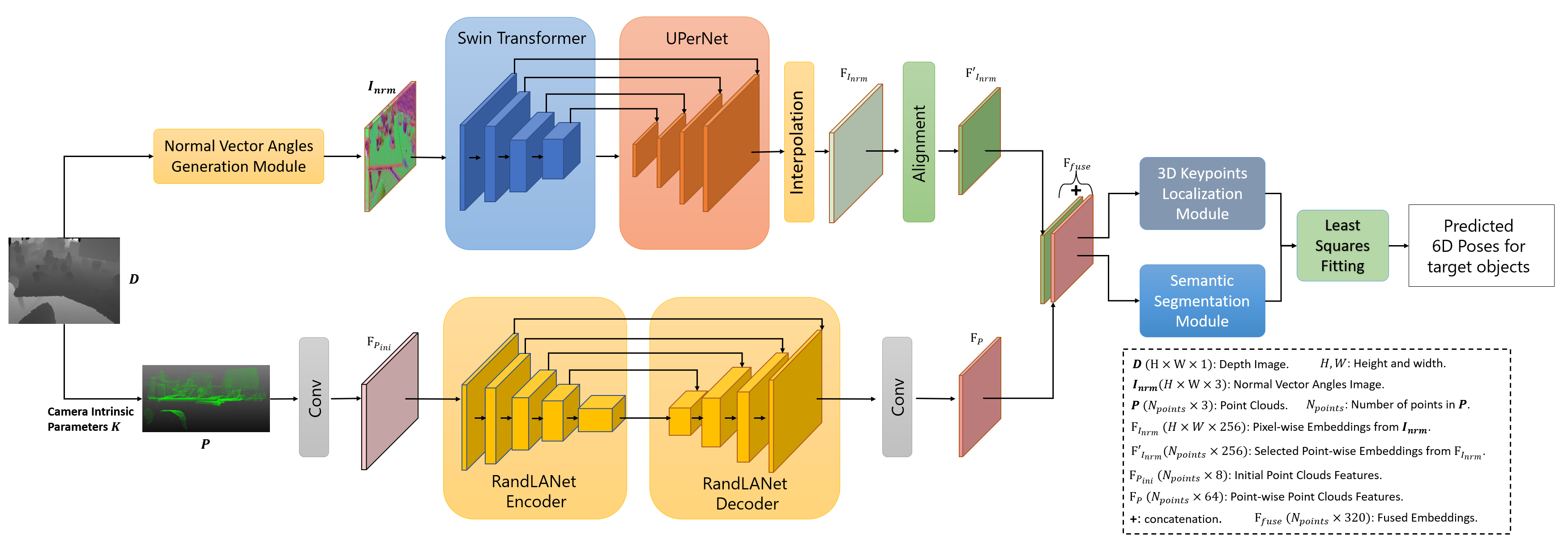}
\caption{The pipeline of our proposed framework. The normal vector angles generation module outputs the normal vector angles image. Two encoder-decoder networks are utilized for representation learning of normal vector angles images and point clouds, respectively. The extracted image and point clouds embeddings are concatenated and then fed into the semantic segmentation and 3D keypoints localization modules to predict the mask and 3D keypoints for the target object. Finally, a least-squares fitting manner is adopted to estimate 6D poses from the predicted 3D keypoints.}
\label{over}
\end{figure*}

\subsection{Overview.} 
Our proposed framework follows a pipeline that consists of several steps, as shown in Fig. \ref{over}. First, we feed the depth image $D$ into the normal vector angles generation module, which produces the normal vector angles image $I_{nrm}$. At the same time, we lift $D$ to point clouds $P$ using the given camera parameters $K$. We then utilize two encoder-decoder networks to learn the representations of $I_{nrm}$ and $P$, respectively. Specifically, the image embeddings $\mathrm{F}_{I_{nrm}}^{\prime}$ and point clouds embeddings $\mathrm{F}_p$ are extracted and concatenated to form the fused features $\mathrm{F}_{fuse}$. We then feed $\mathrm{F}_{fuse}$ into the semantic segmentation and 3D keypoints localization modules to predict the mask and 3D keypoints of the target object. Finally, we adopt a least-squares fitting method to estimate the 6D poses based on the predicted 3D keypoints.

\subsection{Normal Vector Angles Image Generation.}

Previous depth-based 6D pose estimation studies \cite{cai2022ove6d,liu2022catre} directly fed depth information into image encoders, but these approaches may not fully utilize the geometry information in depth images. Depth images capture depth information and 2D grid structures and contain implicit local geometric features and directional information.

\begin{figure}[h] 
\includegraphics[width=0.5\textwidth]{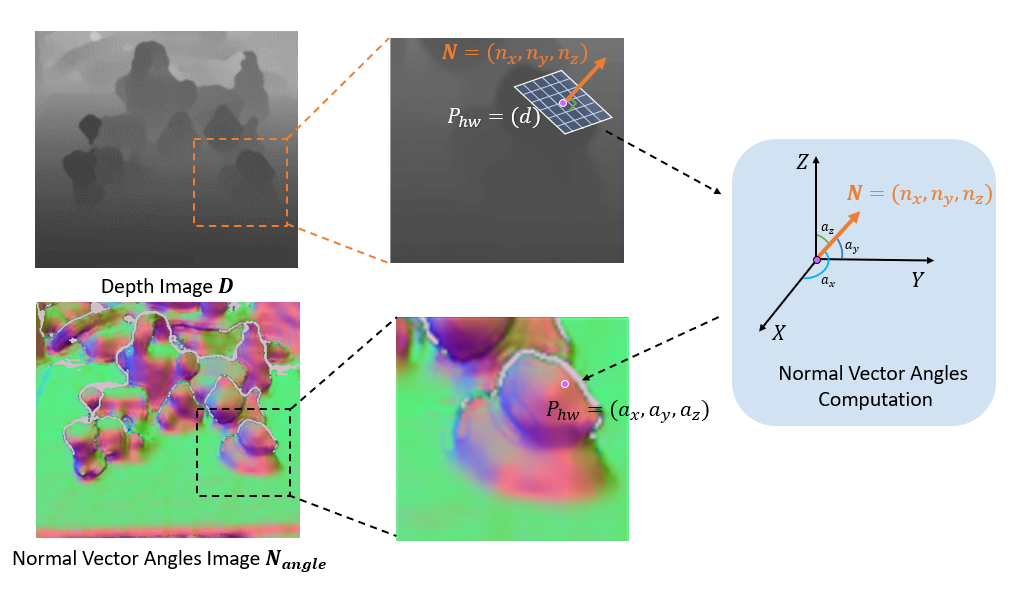}
\caption{Examples of the normal vector angles image generation module. A pixel value of $P_{hw}$ in $D$ is the distance $d$ between an object's surface point and the camera. We compute the normal vector $\mathbf{N}(n_x,n_y,n_z)$ for each surface point based on its depth information, and then obtain the angles $(a_x, a_y, a_z)$ between $\mathbf{N}$ and $XYZ$ axes.}
\label{nor}
\end{figure}

To capture both the local geometric features and directional information from depth images, we generate a \textit{normal vector angles image} for each scene, as shown in Fig. \ref{nor}. To achieve this, we calculate the surface normal vector $\mathbf{N}(n_x, n_y, n_z)$ for each pixel $(u,v)$ in the depth image $D$, and then determine the angles between $\mathbf{N}(n_x, n_y, n_z)$ and the $XYZ$ axes in the camera coordinate system. The $X$-axis, $Y$-axis, and $Z$-axis are represented as vectors $\mathbf{x}(1,0,0)$, $\mathbf{y}(0,1,0)$, and $\mathbf{z}(0,0,1)$, respectively. The angles between $\mathbf{N}(n_x, n_y, n_z)$ and these three axes vectors can be expressed as: 
\begin{equation}
\begin{aligned}
& a_x=\arccos (\mathbf{N} \cdot \mathbf{x}), \\
& a_y=\arccos (\mathbf{N} \cdot \mathbf{y}), \\
& a_z=\arccos (\mathbf{N} \cdot \mathbf{z}).
\end{aligned}
\end{equation}
Finally, the angles $(a_x, a_y, a_z)$ are normalized into the range 0 $\sim$ 255 and used to create the normal vector angles image $I_{nrm}$, where each pixel has the normalized $(a_x,a_y,a_z)$ as its value. $I_{nrm}$ consists of three channels, each representing one of the normalized angles.

\subsection{Image Feature Extraction.}

\begin{figure}[h] 
\includegraphics[width=0.5\textwidth]{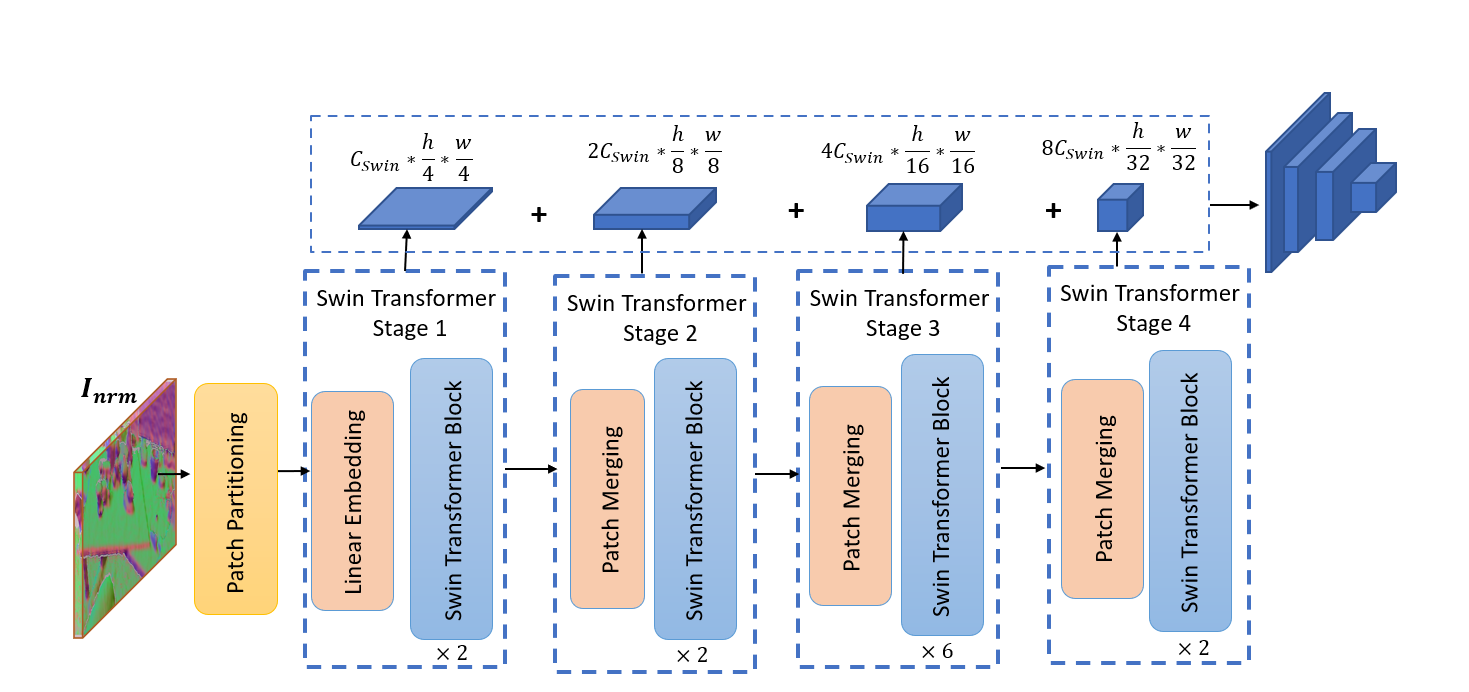}
\caption{The encoder is based on Tiny Swin Transformer (Swin-T), which contains 4 stages. $I_{nrm}$ is fed into the patch partition module, whose outputs are fed into stage 1. $C_{swin}$ represents the embedding dimensions, that $C_{swin}=96$ in Swin-T. $h$ and $w$ are the image height and width. Finally, the embeddings from each stage are stacked together and fed into the following decoder network, as shown in Fig. \ref{over}.}
\label{swin}
\end{figure}

We propose an encoder-decoder network to extract pixel-wise embeddings from normal vector angles image $I_{nrm}$, as shown in the top panel of Fig. \ref{over}. The network employs a Tiny Swin Transformer (Swin-T) \cite{liu2021swin} as the encoder to learn multi-scale representations from $I_{nrm}$. As shown in Fig. \ref{swin}, Swin-T includes 4 stages of Swin Transformer blocks with modified self-attention layers, linear embedding layers, and patch merging layers. $I_{nrm}$ is fed into the encoder as tokens. In stage 1, the linear embedding layer and a Swin Transformer block with modified self-attention layers extract the features. Within stages 2, 3, and 4, patch merging layers reduce feature map resolutions, and Swin Transformer blocks are applied for feature transformation to enlarge feature dimensions. The embeddings from all stages are combined as hierarchical representations and then fed into the decoder network. Compared to the other Swin Transformer models with larger sizes, such as Small, Based, and Large Swin Transformer models, Swin-T has fewer parameters, making it more efficient.

After encoding the normal vector angles image with Swin-T, the network employs UPerNet \cite{xiao2018unified} and bilinear interpolation to generate dense pixel-wise image features. The multi-scale feature maps from Swin-T's are passed through a Pyramid Pooling Module in UPerNet to generate feature maps with the same dimension. These feature maps are resized by a bilinear interpolation process and then concatenated. The dimension of the concatenated features is reduced through a series of convolutional neural networks and an additional interpolation stage, resulting in the pixel-wise image embeddings $\mathrm{F}_{I_{nrm}}$, whose size matches the input image size.

\subsection{Point Clouds Feature Extraction.}

We begin by converting depth images to point clouds $P$ by the given camera parameters. Next, we adopt RandLA-Net \cite{hu2020randla} to extract representations from $P$, as shown in the bottom panel of Fig. \ref{over}. Initially, $P$ goes through a series of convolutional neural networks to generate initial features $\mathrm{F}_{P_{ini}}$. These initial features are then input to RandLA-Net encoder and decoder. The encoder consists of five stages that extract multi-scale embeddings, while the decoder recovers the resolution of the features in each stage. Finally, we apply additional convolutional neural networks to produce the final point-wise point clouds embeddings $\mathrm{F}_P$.

\subsection{3D Keypoint-based Pose Estimation.}
Recently, several works on 6D pose estimation \cite{liu20213d,he2021ffb6d,he2020pvn3d} estimate 6D poses by establishing keypoint correspondences between 3D models and input RGB-D images, and then applying a least-squares fitting algorithm to recover 6D poses based on these correspondences. Similar to \cite{he2021ffb6d,he2020pvn3d}, we follow this way to estimate 6D poses. As shown in Fig. \ref{over}, after obtaining pixel-wise image embeddings $\mathrm{F}_{I_{nrm}}$ and point-wise point clouds embeddings $\mathrm{F}_P$, we take advantage of the alignment between the image and the point clouds, and then select the point-wise image features $\mathrm{F}_{I_{nrm}}^{\prime}$ from $\mathrm{F}_{I_{nrm}}$. Afterward, we concatenate $\mathrm{F}_P$ and $\mathrm{F}_{I_{nrm}}^{\prime}$ to obtain the fused features $\mathrm{F}_{fuse}$, which is fed into the semantic segmentation module and 3D keypoints localization module to predict the mask and detect 3D keypoints of the target object. Finally, we adopt a least-squares fitting manner to estimate the object's pose. The approaches are described in detail as follows:

\subsubsection{3D Keypoints Localization}
Inspired from \cite{he2020pvn3d}, we obtain 3D keypoints of the target object by a keypoint voting module. Specifically, the module takes the concatenated features $F_{fuse}$ as input. Then for each 3D point, the module predicts the offsets from the point itself to the selected 3D keypoints of the target object. Given the predicted offsets, the selected 3D keypoints can be voted using a MeanShift clustering method \cite{comaniciu2002mean}.

\subsubsection{Semantic Segmentation} 
We integrate the instance segmentation module into our pipeline. The module predicts each pixel's label to segment the target object from the scene. Unlike previous works \cite{cai2022ove6d,liu2022catre} that require an external segmentation network, such as Mask R-CNN \cite{he2017mask}, to preprocess the input image before feeding it into the pose estimation network, our integrated approach has several benefits. On the one hand, it makes the framework more comprehensive and streamlined. On the other hand, by forcing the segmentation module to distinguish objects, it helps to extract both global and local features, which in turn benefits the 3D keypoints localization module.

\subsubsection{Least Squares Fitting}
To compute the rotation matrix $R$ and the translation vector $T$, given the 3D keypoints in the object coordinates system $\{p_i\}^{N}_{i=1}$ and the corresponding 3D keypoints in the camera coordinates system $\{c_i\}^{N}_{i=1}$, the lease-squares fitting algorithm \cite{arun1987least} minimizes the squared loss:
\begin{equation}
\begin{aligned}
& L=\sum_{i=1}^N\left\|c_i-\left(R p_i+T\right)\right\|^2.
\end{aligned}
\end{equation}

\section{EXPERIMENTS}
We conduct experiments on two public benchmark datasets including the LineMod \cite{hinterstoisser2011multimodal} and the Occlusion-LineMod \cite{brachmann2014learning}. Compared with state-of-the-art baselines, our proposed {\method} outperforms them on both datasets. Besides, we present ablation studies to demonstrate the effectiveness of the components in {\method}. 
\subsection{Experimental Setup.} 
\textbf{Datasets.}
LineMod dataset is wildly used in 6D pose estimation. It contains RGB-D images and 13 indoor texture-less objects in cluttered scenes. Following \cite{he2021ffb6d}, we split the training and testing sets and generated 20K synthetic depth images for each category in LineMod.       

The Occlusion-LineMod dataset is a subset of LineMod and contains 8 objects in LineMod. Each scene in Occlusion-LineMod has heavy occlusions, making the task more challenging. We follow \cite{he2021ffb6d} to split the training and testing sets in Occlusion-LineMod. Each training set in the Occlusion-LineMod category contains 40K photorealistic depth images synthetic rendered by BlenderProc4BOP \cite{denninger2020blenderproc}.   

\textbf{Evaluation Metrics.}
We use ADD and ADDS metrics to evaluate the models' performance for asymmetric and symmetric objects, following \cite{xiang2018posecnn}. For asymmetric objects, ADD computes the mean distance between two transformed 3D CAD model points using the estimated pose and the ground truth pose, defined as follows: 
$$
\mathrm{ADD}=\frac{1}{m} \sum_{x \in M}\|(R x+T)-(\tilde{R} x+\tilde{T})\|, \eqno{(3)}
$$
where $M$ denotes the set of 3D CAD model points and $m$ is the number of points. $R$ and $T$ are the predicted rotation and translation, respectively. $\tilde{R}$ and $\tilde{T}$ are ground truth rotation and translation. $||.||$ denotes the Euclidean norm. 

For rotational symmetric objects, we compute the ADDS, which is the mean distance based on the closest point distance between two transformed 3D CAD model points: 
$$
\mathrm{ADDS}=\frac{1}{m} \sum_{x_{1} \in M} \min _{x_{2} \in M}\left\|\left(R x_{1}+T\right)-\left(\tilde{R} x_{2}+\tilde{T}\right)\right\|. \eqno{(4)}
$$

In our experiments, we report the accuracy in terms of ADD or ADDS less than 10\% of the object diameter as in \cite{he2021ffb6d,hinterstoisser2011multimodal}. 

\subsection{Implementation Details.}

\textbf{Network Architecture.} {\method} uses two encoder-decoder structures to extract features from normal vector angles images and point clouds. For image feature extraction, Swin-T \cite{liu2021swin} is utilized as the encoder to produce feature maps $F_{swin}$ at 4 different resolutions and dimensions. These feature maps have dimensions of 96, 192, 384, and 768 and resolutions of $\frac{H}{4} * \frac{W}{4}$, $\frac{H}{8} * \frac{W}{8}$, $\frac{H}{16} * \frac{W}{16}$, and $\frac{H}{32} * \frac{W}{32}$, respectively, where $H$ and $W$ represent the height and width of the input image. Then, UPerNet decodes and interpolates the multi-scale feature map $F_{swin}$ to the feature map $F_{intep}$ with an identical dimension and resolution: 256 and $\frac{H}{4} * \frac{W}{4}$, respectively. After concatenating all features from $F_{intep}$, a series of convolutional neural networks are applied to reduce the dimension of the concatenated features. Finally, an additional interpolation process expands $F_{concat}$ to $256*H*W$, which are the dense pixel-wise image embeddings $\mathrm{F}_{I_{nrm}}$. For point cloud feature extraction, we first convert depth images to point clouds and sample them to 12288 points following \cite{he2021ffb6d}. We then employ RandLANet to extract point-wise features from the point clouds, with a dimension of 64. With the aforementioned settings, the number of parameters of our model is about 37M.

\textbf{Optimization Regularization.} We apply focal loss \cite{lin2017focal} and L1 loss \cite{he2020pvn3d} to supervise the semantic segmentation module and the 3D keypoints localization module, respectively. We sum the focal loss and L1 loss with various weights to jointly optimize them.

\textbf{Keypoint Detection.} We conduct the SIFT-FPS keypoints selection algorithm from \cite{he2021ffb6d} by detecting 2D keypoints in RGB images using SIFT, then projecting 2D keypoints to 3D in the object coordinates system, and finally applying the FPS algorithm to choose N points from them.

\subsection{Comparison with State-of-the-Art Methods.}
\textbf{Evaluation on the LineMod Dataset.} Our proposed {\method} has been evaluated against state-of-the-art 6D pose estimation methods on the LineMod dataset. We categorize the selected methods into three groups based on their input types: RGB, RGB-D, and Depth-Only. Table \ref{tab:Table 1} presents the comparison of the ADD(S) accuracy of {\method} and other methods without any post-processing refinement. We compute ADD for non-symmetric objects and ADDS for symmetric objects. The results demonstrate that without the ground-truth (GT) mask, our {\method} outperforms the previous best-performed depth-based baseline, CATRE \cite{liu2022catre} by 5.94\% on the accuracy of ADD(S) metric. With the GT masks, our {\method} still outperforms the previous best baseline, OVE6D (with GT masks) by 1.14\%.

\begin{figure*} 
\includegraphics[width=\textwidth]{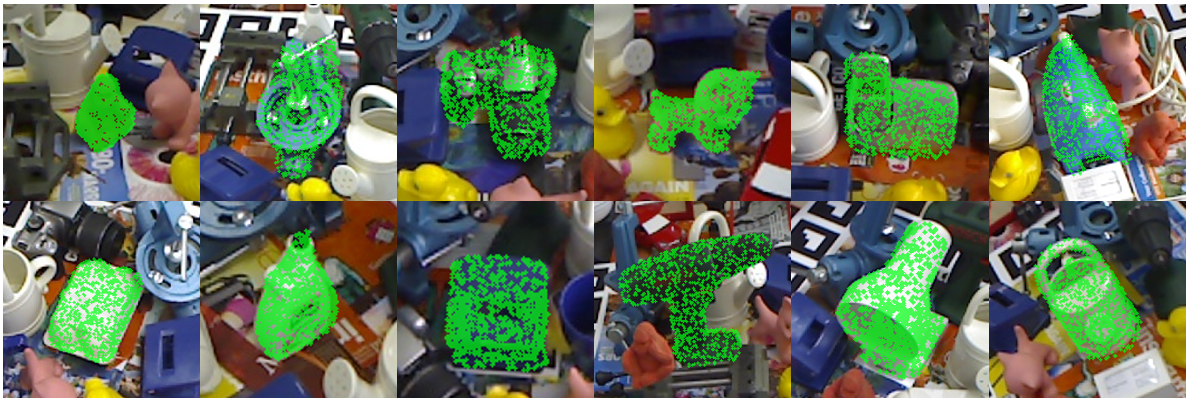}
\caption{For the qualitative evaluation of our proposed {\method} on the LineMod dataset, we transform the 3D surface points from the object meshes using the predicted pose to the camera coordinate system and then project them onto the image using the camera intrinsic matrix. To better visualize the results, we use RGB images to display our results, with the projected points shown in green.}
\label{LM_Quali}
\end{figure*}

Moreover, our proposed method outperforms some RGB-based or RGB-D based methods on certain categories, even without GT masks and only with geometry information as input. This highlights the effectiveness of our method in leveraging the geometry information from depth images for 6D object pose estimation. Qualitative results showing the performance of {\method} are displayed in Fig. \ref{LM_Quali}, which demonstrate that our proposed model can accurately predict 6D object poses.

\begin{table*}[]
\centering
\caption{The accuracy in terms of ADD(S) results for the LineMod dataset. Symmetric objects are noted with *. We highlight the best performance in bold for each group.}
\label{tab:Table 1}
\resizebox{\textwidth}{!}{%
\begin{tabular}{c||ccc||ccc||cccccc}
\toprule
INPUTS & \multicolumn{3}{c||}{RGB} & \multicolumn{3}{c||}{RGB-D} & \multicolumn{6}{c}{Depth-Only} \\ \midrule
METHODS & \multicolumn{1}{c|}{PVNet \cite{peng2019pvnet}} & \multicolumn{1}{c|}{Pix2Pose \cite{park2019pix2pose}} & RNNPose \cite{xu2022rnnpose} & \multicolumn{1}{c|}{PVN3D \cite{he2020pvn3d}} & \multicolumn{1}{c|}{DenseFusion \cite{wang2019densefusion}} & KPD \cite{saadi2021optimizing} & \multicolumn{1}{c|}{CloudAAE \cite{gao2021cloudaae}} & \multicolumn{1}{c|}{\cellcolor[HTML]{FFFFFF}CATRE \cite{liu2022catre}} & \multicolumn{1}{c|}{\begin{tabular}[c]{@{}c@{}}OVE6D \cite{cai2022ove6d}\\ (w Mask R-CNN)\end{tabular}} & \multicolumn{1}{c|}{\begin{tabular}[c]{@{}c@{}}OVE6D \cite{cai2022ove6d}\\ (w GT Masks)\end{tabular}} & \multicolumn{1}{c|}{\begin{tabular}[c]{@{}c@{}}OURS\\ (w/o GT Mask)\end{tabular}} & \begin{tabular}[c]{@{}c@{}}OURS\\ (w GT Masks)\end{tabular} \\ \midrule
ape & \multicolumn{1}{c|}{43.6} & \multicolumn{1}{c|}{58.1} & \textbf{88.2} & \multicolumn{1}{c|}{\textbf{97.3}} & \multicolumn{1}{c|}{92} & 94.2 & \multicolumn{1}{c|}{74.5} & \multicolumn{1}{c|}{63.7} & \multicolumn{1}{c|}{-} & \multicolumn{1}{c|}{-} &\multicolumn{1}{c|}{91.7} & \textbf{95.4} \\
benchvise & \multicolumn{1}{c|}{\textbf{99.9}} & \multicolumn{1}{c|}{91.0} & 79.7 & \multicolumn{1}{c|}{\textbf{99.7}} & \multicolumn{1}{c|}{93} & 98.2 & \multicolumn{1}{c|}{86.6} & \multicolumn{1}{c|}{\textbf{98.6}} & \multicolumn{1}{c|}{-} & \multicolumn{1}{c|}{-} &\multicolumn{1}{c|}{97.9} & 98.2 \\
camera & \multicolumn{1}{c|}{86.9} & \multicolumn{1}{c|}{60.9} & \textbf{98.0} & \multicolumn{1}{c|}{\textbf{99.6}} & \multicolumn{1}{c|}{94} & 98.5 & \multicolumn{1}{c|}{65.6} & \multicolumn{1}{c|}{89.7} & \multicolumn{1}{c|}{-} & \multicolumn{1}{c|}{-} &\multicolumn{1}{c|}{94.8} & \textbf{96.9} \\
can & \multicolumn{1}{c|}{95.5} & \multicolumn{1}{c|}{84.4} & \textbf{99.3} & \multicolumn{1}{c|}{\textbf{99.5}} & \multicolumn{1}{c|}{93} & 94.0 & \multicolumn{1}{c|}{90.2} & \multicolumn{1}{c|}{96.1} & \multicolumn{1}{c|}{-} & \multicolumn{1}{c|}{-} &\multicolumn{1}{c|}{97.6} & \textbf{98.2} \\
cat & \multicolumn{1}{c|}{79.3} & \multicolumn{1}{c|}{65.0} & \textbf{96.4} & \multicolumn{1}{c|}{\textbf{99.8}} & \multicolumn{1}{c|}{97} & 92.0 & \multicolumn{1}{c|}{90.7} & \multicolumn{1}{c|}{84.3} & \multicolumn{1}{c|}{-} & \multicolumn{1}{c|}{-} &\multicolumn{1}{c|}{98.3} & \textbf{98.6} \\
driller & \multicolumn{1}{c|}{96.4} & \multicolumn{1}{c|}{76.3} & \textbf{99.7} & \multicolumn{1}{c|}{\textbf{99.3}} & \multicolumn{1}{c|}{87} & 97.2 & \multicolumn{1}{c|}{97.3} & \multicolumn{1}{c|}{\textbf{98.6}} & \multicolumn{1}{c|}{-} & \multicolumn{1}{c|}{-} &\multicolumn{1}{c|}{\textbf{98.6}} & 98.5 \\
duck & \multicolumn{1}{c|}{52.6} & \multicolumn{1}{c|}{43.8} & \textbf{89.3} & \multicolumn{1}{c|}{\textbf{98.2}} & \multicolumn{1}{c|}{92} & 91.5 & \multicolumn{1}{c|}{50.0} & \multicolumn{1}{c|}{63.9} & \multicolumn{1}{c|}{-} & \multicolumn{1}{c|}{-} &\multicolumn{1}{c|}{88.5} & \textbf{92.7} \\
eggbox* & \multicolumn{1}{c|}{99.2} & \multicolumn{1}{c|}{96.8} & \textbf{99.5} & \multicolumn{1}{c|}{99.8} & \multicolumn{1}{c|}{\textbf{100}} & 99.6 & \multicolumn{1}{c|}{99.7} & \multicolumn{1}{c|}{99.8} & \multicolumn{1}{c|}{-} & \multicolumn{1}{c|}{-} &\multicolumn{1}{c|}{\textbf{100.0}} & \textbf{100.0} \\
glue* & \multicolumn{1}{c|}{95.7} & \multicolumn{1}{c|}{79.4} & \textbf{99.7} & \multicolumn{1}{c|}{\textbf{100.0}} & \multicolumn{1}{c|}{\textbf{100}} & 92.5 & \multicolumn{1}{c|}{93.5} & \multicolumn{1}{c|}{99.4} & \multicolumn{1}{c|}{-} & \multicolumn{1}{c|}{-} &\multicolumn{1}{c|}{98.6} & \textbf{100.0} \\
holepuncher & \multicolumn{1}{c|}{82.0} & \multicolumn{1}{c|}{74.8} & \textbf{97.4} & \multicolumn{1}{c|}{\textbf{99.9}} & \multicolumn{1}{c|}{92} & 92.1 & \multicolumn{1}{c|}{57.9} & \multicolumn{1}{c|}{93.2} & \multicolumn{1}{c|}{-} & \multicolumn{1}{c|}{-} &\multicolumn{1}{c|}{92.4} & \textbf{93.6} \\
iron & \multicolumn{1}{c|}{98.8} & \multicolumn{1}{c|}{83.1} & \textbf{100.0} & \multicolumn{1}{c|}{\textbf{99.7}} & \multicolumn{1}{c|}{97} & 98.7 & \multicolumn{1}{c|}{85.0} & \multicolumn{1}{c|}{\textbf{98.4}} & \multicolumn{1}{c|}{-} & \multicolumn{1}{c|}{-} &\multicolumn{1}{c|}{96.9} & 96.9 \\
lamp & \multicolumn{1}{c|}{99.3} & \multicolumn{1}{c|}{82.0} & \textbf{99.8} & \multicolumn{1}{c|}{\textbf{99.8}} & \multicolumn{1}{c|}{95} & 96.5 & \multicolumn{1}{c|}{82.1} & \multicolumn{1}{c|}{98.7} & \multicolumn{1}{c|}{-} & \multicolumn{1}{c|}{-} &\multicolumn{1}{c|}{98.8} & \textbf{99.1} \\
phone & \multicolumn{1}{c|}{92.4} & \multicolumn{1}{c|}{45.0} & \textbf{98.4} & \multicolumn{1}{c|}{\textbf{99.5}} & \multicolumn{1}{c|}{93} & 97.2 & \multicolumn{1}{c|}{94.4} & \multicolumn{1}{c|}{97.5} & \multicolumn{1}{c|}{-} & \multicolumn{1}{c|}{-} &\multicolumn{1}{c|}{98.3} & \textbf{98.8} \\ \midrule
MEAN & \multicolumn{1}{c|}{86.3} & \multicolumn{1}{c|}{72.4} & \textbf{97.4} & \multicolumn{1}{c|}{\textbf{99.4}} & \multicolumn{1}{c|}{94} & 95.6 & \multicolumn{1}{c|}{82.1} & \multicolumn{1}{c|}{90.9} & \multicolumn{1}{c|}{86.1} & \multicolumn{1}{c|}{96.4} &\multicolumn{1}{c|}{96.3} & \textbf{97.5} \\ \bottomrule
\end{tabular}%
}
\end{table*}

\begin{table*}[]
\centering
\caption{The accuracy in terms of ADD(S) results for the Occlusion LineMod dataset. Symmetric objects are noted with *. We highlight the best performance in bold for each group.}
\label{tab:Table 2}
\resizebox{\textwidth}{!}{%
\begin{tabular}{@{}c||ccc||ccc||ccccc@{}}
\toprule
INPUTS & \multicolumn{3}{c||}{RGB} & \multicolumn{3}{c||}{RGB-D} & \multicolumn{5}{c}{Depth-Only} \\ \midrule
METHODS & \multicolumn{1}{c|}{PVNet \cite{peng2019pvnet}} & \multicolumn{1}{c|}{Pix2Pose \cite{park2019pix2pose}} & Keypoint \cite{zhang2021keypoint} & \multicolumn{1}{c|}{Point-to-Keypoint \cite{hua20203d}} & \multicolumn{1}{c|}{FFB6D \cite{he2021ffb6d}} & KPD \cite{saadi2021optimizing} & \multicolumn{1}{c|}{CloudAAE \cite{gao2021cloudaae}} & \multicolumn{1}{c|}{\begin{tabular}[c]{@{}c@{}}OVE6D \cite{cai2022ove6d}\\ (w Mask R-CNN)\end{tabular}} & \multicolumn{1}{c|}{\begin{tabular}[c]{@{}c@{}}OVE6D \cite{cai2022ove6d}\\ (w GT Masks)\end{tabular}} & \multicolumn{1}{c|}{\begin{tabular}[c]{@{}c@{}}OURS\\ (w/o GT Mask)\end{tabular}} & \begin{tabular}[c]{@{}c@{}}OURS\\ (w GT Masks)\end{tabular} \\ \midrule
ape & \multicolumn{1}{c|}{15.8} & \multicolumn{1}{c|}{\textbf{22.0}} & - & \multicolumn{1}{c|}{\textbf{51.6}} & \multicolumn{1}{c|}{47.2} & 19.5 & \multicolumn{1}{c|}{-} & \multicolumn{1}{c|}{-} & \multicolumn{1}{c|}{-} &\multicolumn{1}{c|}{50.3} & \textbf{59.8} \\
can & \multicolumn{1}{c|}{\textbf{63.3}} & \multicolumn{1}{c|}{44.7} & - & \multicolumn{1}{c|}{75.6} & \multicolumn{1}{c|}{\textbf{85.2}} & 78.4 & \multicolumn{1}{c|}{-} & \multicolumn{1}{c|}{-} & \multicolumn{1}{c|}{-} & \multicolumn{1}{c|}{84.1} & \textbf{88.8} \\
cat & \multicolumn{1}{c|}{16.7} & \multicolumn{1}{c|}{\textbf{22.7}} & - & \multicolumn{1}{c|}{28.7} & \multicolumn{1}{c|}{\textbf{45.7}} & 28.2 & \multicolumn{1}{c|}{-} & \multicolumn{1}{c|}{-} & \multicolumn{1}{c|}{-} & \multicolumn{1}{c|}{39.0} & \textbf{46.7} \\
driller & \multicolumn{1}{c|}{25.2} & \multicolumn{1}{c|}{\textbf{44.7}} & - & \multicolumn{1}{c|}{66.9} & \multicolumn{1}{c|}{\textbf{81.4}} & 75.1 & \multicolumn{1}{c|}{-} & \multicolumn{1}{c|}{-} & \multicolumn{1}{c|}{-} & \multicolumn{1}{c|}{88.9} & \textbf{95.1} \\
duck & \multicolumn{1}{c|}{\textbf{65.7}} & \multicolumn{1}{c|}{15.0} & - & \multicolumn{1}{c|}{36.7} & \multicolumn{1}{c|}{\textbf{53.9}} & 38.6 & \multicolumn{1}{c|}{-} & \multicolumn{1}{c|}{-} & \multicolumn{1}{c|}{-} &\multicolumn{1}{c|}{53.0} & \textbf{59.4} \\
eggbox* & \multicolumn{1}{c|}{\textbf{50.2}} & \multicolumn{1}{c|}{25.2} & - & \multicolumn{1}{c|}{47.1} & \multicolumn{1}{c|}{\textbf{70.2}} & 51.2 & \multicolumn{1}{c|}{-} & \multicolumn{1}{c|}{-} & \multicolumn{1}{c|}{-} &\multicolumn{1}{c|}{28.5} & \textbf{90.3} \\
glue* & \multicolumn{1}{c|}{\textbf{49.6}} & \multicolumn{1}{c|}{32.4} & - & \multicolumn{1}{c|}{\textbf{71.9}} & \multicolumn{1}{c|}{60.1} & 52.1 & \multicolumn{1}{c|}{-} & \multicolumn{1}{c|}{-} & \multicolumn{1}{c|}{-} &\multicolumn{1}{c|}{58.4} & \textbf{88.0} \\
holepuncher & \multicolumn{1}{c|}{39.7} & \multicolumn{1}{c|}{\textbf{49.5}} & - & \multicolumn{1}{c|}{45.7} & \multicolumn{1}{c|}{\textbf{85.9}} & 59.0 & \multicolumn{1}{c|}{-} & \multicolumn{1}{c|}{-} & \multicolumn{1}{c|}{-} &\multicolumn{1}{c|}{79.8} & \textbf{88.8} \\ \midrule
MEAN & \multicolumn{1}{c|}{\textbf{40.8}} & \multicolumn{1}{c|}{32.0} & 33.7 & \multicolumn{1}{c|}{52.6} & \multicolumn{1}{c|}{\textbf{66.2}} & 50.3 & \multicolumn{1}{c|}{58.9} & \multicolumn{1}{c|}{56.1} & \multicolumn{1}{c|}{70.9} &\multicolumn{1}{c|}{60.3} & \textbf{77.1} \\ \bottomrule
\end{tabular}%
}

\end{table*}

\begin{figure*}
\includegraphics[width=\textwidth]{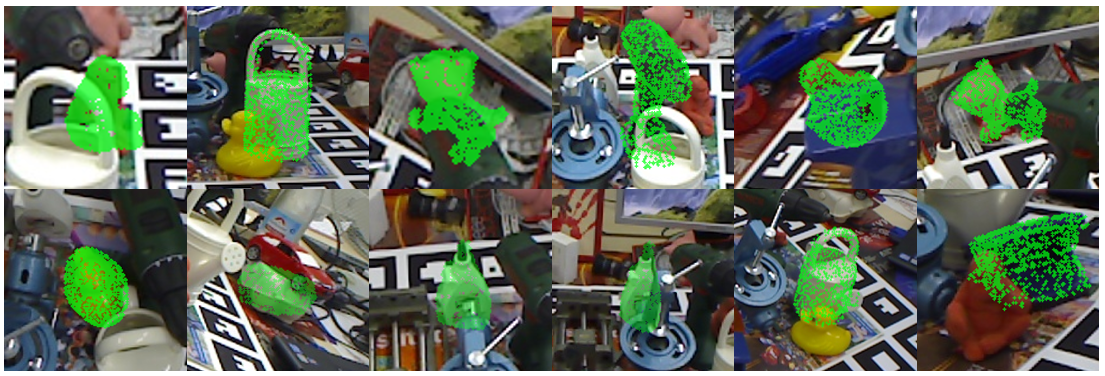}
\caption{Qualitative results of the Occlusion LineMod Dataset. We project 3D surface points from object meshes to the image plane using the predicted pose and the camera intrinsic matrix. We use the RGB image to visualize the projected points shown in green. }
\label{tab:Occ}
\end{figure*}

\textbf{Evaluation on the Occlusion LineMod Dataset.} For the Occlusion LineMod dataset, the results in Table \ref{tab:Table 2} indicate that {\method} outperforms the best-performing depth-based baseline, which is CloudAAE, by 2.38\% without GT masks. With GT masks, {\method} achieves an 8.74\% higher accuracy on ADD(S) compared to OVE6D (with GT masks). This improvement is observed without any post-processing refinement, highlighting the robustness of our model. Furthermore, {\method} exceeds some RGB-based or RGB-D based methods on certain categories even without GT masks. Due to heavy occlusions, GT masks significantly improve our performance since our semantic segmentation module faces difficulty in segmenting objects. Qualitative results are displayed in Fig. \ref{tab:Occ}, which show accurate 6D pose estimation results even in scenes with heavy occlusion, as demonstrated in the images.

\subsection{Ablation Study.}
\begin{table}[]
\centering
\caption{Results of ablation study. We validate the efficacy of different aspects of SwinDePose, The numbers with blue color indicate the performance difference between our full model and the corresponding ablated model. }
\label{tab:Table 3}
\resizebox{0.5\textwidth}{!}{%
\begin{tabular}{@{}|c|c|rr|ll|@{}}
\toprule
Aspect & Description & \multicolumn{2}{c|}{\begin{tabular}[c]{@{}c@{}}Average ADD(S) \\ w/o GT Masks\end{tabular}} & \multicolumn{2}{c|}{\cellcolor[HTML]{FFFFFF}\begin{tabular}[c]{@{}c@{}}Average ADD(S) \\ w GT Masks\end{tabular}} \\ \midrule
\midrule
Full Model & \begin{tabular}[c]{@{}c@{}}SwinDePose\end{tabular} & \multicolumn{2}{c|}{96.3} & \multicolumn{2}{c|}{97.5} \\ \midrule
\midrule
 & w/o Any Encoder & {\color[HTML]{333333}  \ \ 42.8} & \textcolor{blue}{(-53.5)} \ \ & \ \ 51.6 & \textcolor{blue}{(-45.9)} \ \ \\ \cmidrule(l){2-6} 
\multirow{-2}{*}{\begin{tabular}[c]{@{}c@{}}Images Feature Extraction\end{tabular}} & w Conv Encoder & \ \ 94.3 & \textcolor{blue}{(-2.0)} \ \ & \ \ 95.7 & \ \textcolor{blue}{(-1.8)} \\ \midrule
\midrule
 & w/o Any Encoder & 51.4 & \textcolor{blue}{(-44.9)} \ \  & \ \ 59.3 & \textcolor{blue}{(-38.2)} \ \\ \cmidrule(l){2-6} 
\multirow{-2}{*}{\begin{tabular}[c]{@{}c@{}}Point Clouds \\ Feature Extraction\end{tabular}} & w PointNet & \ \ 84.0 & \textcolor{blue}{(-12.3)} \ \ & \ \ 90.1 & \ \textcolor{blue}{(-7.4)} \\ \midrule
\midrule
 & w Depth Images & 94.1 & \textcolor{blue}{(-2.2)} \ \ & \ \ 95.7 & \ \textcolor{blue}{(-1.8)} \\ \cmidrule(l){2-6} 
\multirow{-2}{*}{\begin{tabular}[c]{@{}c@{}}Normal Vector Angles \\ Images Replacement\end{tabular}} & w Normal Vectors & 81.2 & \textcolor{blue}{(-15.1)} \ \ & \ \ 86.2 &  \textcolor{blue}{(-11.3)} \\ \bottomrule
\end{tabular}%
}
\end{table}

In this subsection, we display extensive ablation studies on our model design and discuss their effects on the LineMod dataset. The results shown in Table \ref{tab:Table 3} are the average ADD(S) on the LineMod dataset. 

\textbf{Effect of the Images Feature Extraction.} 
To validate the effectiveness of using Swin-T for learning normal vector angles image embeddings, we conducted an ablation study in Table \ref{tab:Table 3}. We tested two scenarios: without applying any module to learn representations from images (w/o Any Encoder), and using a 4-layer convolutional neural network (w Conv Encoder). The results in Table \ref{tab:Table 3} indicate that it is challenging to estimate accurate 6D poses based solely on point clouds embeddings. Furthermore, when compared to the results obtained using a 4-layer convolutional network, Swin-T architecture achieved the better pose estimation outcomes, highlighting the superiority of using Swin Transformer module for learning image representations.

\textbf{Effect of the Point Clouds Feature Extraction.} 
In this section, we evaluate the impact of various point clouds representation learning modules. We present the results in Table \ref{tab:Table 3}, where we compare two scenarios: one without using any modules for point clouds embeddings learning (w/o Any Encoder), and the other one using PointNet \cite{qi2017pointnet} (w PointNet) to extract point clouds features. Our results show that removing the point cloud embeddings significantly harms the performance. Additionally, we observe that replacing RandLA-Net with PointNet \cite{qi2017pointnet} leads to a decline in the estimation accuracy. We hypothesize that the randomized aggregation method in RandLA-Net allows for more robust feature learning and is less sensitive to individual point errors than PointNet.

\textbf{Effect of the Normal Vector Angles Image.}
To validate the effectiveness of the normal vector angles images, we conducted experiments by replacing them with depth images and replacing them with normal vectors. The results of these experiments are shown in Table \ref{tab:Table 3}. The experimental results indicate that neither using depth images nor normal vectors surpasses using normal vector angles images, demonstrating that the usage of normal vector angles images is beneficial in improving pose estimation accuracy.

\section{CONCLUSIONS}
We introduce {\method}, a novel fusion network for learning representations from a single depth image that maximizes the information present in the scene for 6D pose estimation. In addition, we develop an effective module that converts depth images into normal vector angles images, explicitly incorporating more geometric information into the fusion network. Compared to state-of-the-art methods, our approach achieves superior results on the LineMod and Occlusion-LineMod benchmark datasets. Furthermore, our proposed fusion network can be applied to 6D pose estimation based on RGB-D images, and we anticipate further research in this area.

\section*{Acknowledgement}
This work was partially supported by NSF Award CNS1625843.

\addtolength{\textheight}{0cm}   





\bibliographystyle{unsrt}
\bibliography{IEEEexample}

\end{document}